\documentclass[letterpaper, 10 pt, conference]{ieeeconf}  %

\IEEEoverridecommandlockouts                              %

\usepackage{graphicx}
\usepackage{amsmath}
\usepackage{todonotes}
\usepackage{cleveref}
\usepackage{tabularx}
\usepackage{outlines}
\usepackage{glossaries}

\usepackage{tikz}
\usetikzlibrary{backgrounds}
\usetikzlibrary{fit}
\usetikzlibrary{positioning} %
\usetikzlibrary{calc}
\usetikzlibrary{patterns} %
\usetikzlibrary{quotes} %
\usetikzlibrary{calc}
\usetikzlibrary{intersections}
\usepackage{caption}

\newacronym{hd}{HD}{Hausdorff Distance}
\newacronym{ct}{CT}{Computed Tomography}
\newacronym{rapn}{RAPN}{Robot-Assisted Partial Nephrectomy}
\newacronym{ras}{RAS}{Robotic Assisted Surgery}
\newacronym{cpd}{CPD}{Coherent Point Drift}
\newacronym{hu}{HU}{Hounsfield Unit}
\newacronym{dap}{DAP}{Dose Area Product}
\newacronym{iou}{IoU}{Intersection over Union}
\newacronym{miou}{mIoU}{Mean Intersection over Union}
\newacronym{dpp}{SPC}{Sensor Point Cloud}
\newacronym{mlp}{MLP}{Multilayer Perceptron}

\title{\LARGE \bf
Tracking Tumors under Deformation from Partial Point Clouds using Occupancy Networks
}

\author{Pit Henrich$^{1\dagger}$, Jiawei Liu$^{2\dagger}$, Jiawei Ge$^{2}$, Samuel Schmidgall$^{2}$, Lauren Shepard$^{3}$,\\ Ahmed Ezzat Ghazi$^{3}$, Franziska Mathis-Ullrich$^{1}$, Axel Krieger$^{2}$%
\thanks{\textbf{This work has been submitted to the IEEE for possible publication. Copyright may be transferred without notice, after which this version may no longer be accessible.}}
\thanks{*Research reported in this paper was supported by the Advanced Research Projects Agency for Health (ARPA-H) under grant number AY1AX000023, and National Science Foundation (NSF/FRR CAREER) under grant number 2144348. 
The content is solely the responsibility of the authors and does not necessarily represent the official views of the Advanced Research Projects Agency for Health or the National Science Foundation.}
\thanks{$^{\dagger}$These authors contributed equally.}
\thanks{$^{1}$P. Henrich and F. Mathis-Ullrich are with the Department Artificial Intelligence in Biomedical Engineering (AIBE), Friedrich-Alexander University Erlangen-Nürnberg
(FAU), 91052 Erlangen, Germany. 
        {\tt\small \{pit.henrich, franziska.mathis-ullrich\}@fau.de}}%
\thanks{$^{2}$ J. Liu, J. Ge, S. Schmidgall and A. Krieger are with the Department of Mechanical
Engineering, Johns Hopkins University, Baltimore, MD 21218, USA
        {\tt\small \{jliu298, jge9, sschmi46, axel\}@jhu.edu}}%
\thanks{$^{3}$ L. Shepard and A.E. Ghazi are with the Department of Urology, Johns Hopkins University, Baltimore, MD 21218, USA
        {\tt\small \{lshepar9, aghazi1\}@jh.edu}}%
}
\begin{document}

\maketitle
\thispagestyle{empty}
\pagestyle{empty}

\begin{abstract}
To track tumors during surgery, information from preoperative CT scans is used to determine their position. However, as the surgeon operates, the tumor may be deformed which presents a major hurdle for accurately resecting the tumor, and can lead to surgical inaccuracy, increased operation time, and excessive margins. This issue is particularly pronounced in robot-assisted partial nephrectomy (RAPN), where the kidney undergoes significant deformations during operation. Toward addressing this, we introduce a occupancy network-based method for the localization of tumors within kidney phantoms undergoing deformations at interactive speeds.
We validate our method by introducing a 3D hydrogel kidney phantom embedded with exophytic and endophytic renal tumors.
It closely mimics real tissue mechanics to simulate kidney deformation during in vivo surgery, providing excellent contrast and clear delineation of tumor margins to enable automatic threshold-based segmentation.
Our findings indicate that the proposed method can localize tumors in moderately deforming kidneys with a margin of 6mm to 10mm, while providing essential volumetric 3D information at over 60Hz.
This capability directly enables downstream tasks such as robotic resection.
\end{abstract}

\section{Introduction}

Kidney cancer is one of the most common forms of cancer in the US, with over 65,000 new patients being diagnosed 
every year, leading to over 15,000 deaths~\cite{Cancer_statistics}. 
The standard treatment for localized small renal masses has shifted from radical nephrectomy (complete kidney removal) toward the more minimally invasive approach of partial nephrectomy (removal of the tumor, retaining partial kidney function). 
One of the main challenges during tumor removal is ensuring the resection of adequate tumor margins.
The margin needs to ensure that no cancer cells remain in the kidney, while also avoiding excessive removal of healthy tissue to preserve organ function.

For operations involving tumor resection, surgeons commonly interpret 2D preoperative scans and mentally construct a 3D anatomical model during the surgery.
This process of creating a mental model can extend the duration of the procedure, increasing the risk of damaging adjacent tissues~\cite{cannon20243d}.
The difficulty is exacerbated by tissue deformation during surgery, causing the internalized mental model to become inaccurate over time.

Registration methods that overlay pre-operative data onto the intra-operative scene can provide surgeons with valuable location information for regions of interest.
Altamar et al.~\cite{Altamar_Deformation} develop a deformable registration method based on biomechanical elastic models to simulate in-vivo deformation during partial nephrectomy.
While their method accurately analyzes kidney deformation, it doesn't address the tumor and margin deformation.

\begin{figure}
    \centering
    \begin{tikzpicture}[node distance=0.3cm, auto]

        \node (Phantom) [, rectangle, align=center] {
            \includegraphics[height=2cm]{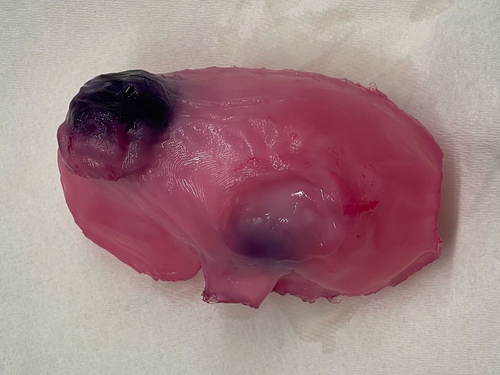}\includegraphics[height=2cm,angle=90]{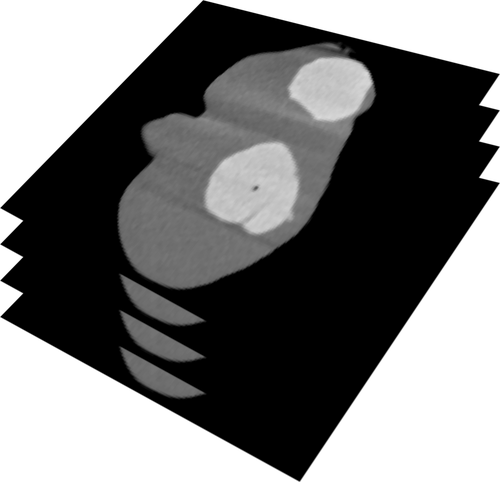}\\Kidney Phantom
        };
        \node (Sensor Point Cloud) [, rectangle, align=center, below=of Phantom] {
            \includegraphics[width=2cm]{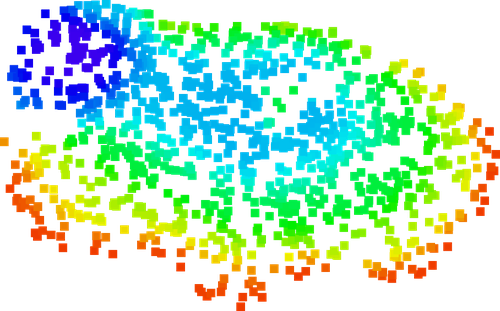}\\Sensor Point Cloud
        };
        \node (Occupancy Point Cloud) [, rectangle, align=center, below=of Sensor Point Cloud] {
            \includegraphics[width=2cm]{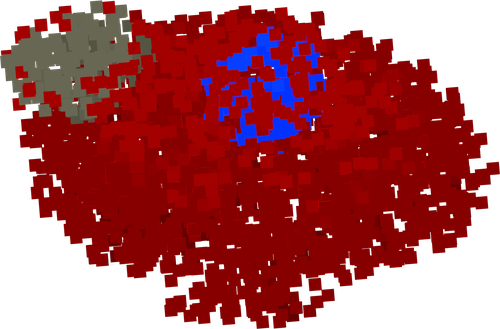}\\Occupancy Point Cloud
        };

            \draw[->] (Phantom) -- (Sensor Point Cloud);
            \draw[->] (Sensor Point Cloud) -- (Occupancy Point Cloud);
     
    \end{tikzpicture}
    \caption{Given a known kidney phantom with two tumors and a pre-operative CT scan, we estimate, from a single sensor point cloud derived from a depth image, a dense occupancy point cloud that encodes the locations of each tumor.}
    \label{fig:figure_one}
\end{figure}

Point-based registration enables the calculation of precise transformations between markers in pre-operative images and those in the physical space, as shown on intra-operative images.
In the absence of distinct anatomical landmarks on organs like the kidney, applying fiducials to the renal surface is advocated for point-based registration~\cite{ge2019landmark, simpfendorfer2016intraoperative}.
Nimmagadda et al.~\cite{nimmagadda2022patient} develop a touch-based registration approach using ink fiducials tattooed on the kidney surface.
They employ the Iterative Closest Point (ICP) algorithm for initial alignment with a pre-operative point cloud, followed by point-based re-registration to account for kidney deformation.
While precise, tattooing intra-operative tissue has been associated with increased patient morbidity~\cite{cannon20243d}.
Zhang et al.~\cite{Xiaohui_AR_registration} propose a different approach. They gather surface data by capturing multiple views of the kidney, stitching them together to construct a 3D surface.
They then apply \gls{cpd} for deformable registration.
This technique achieves markerless deformable registration but requires multiple views of the organ, wherein the organ's deformation during the image collection process could introduce errors.

In previous work~\cite{Henrich_2024_WACV}, we demonstrate a method for reconstructing deformed objects composed of multiple parts from single-viewpoint point clouds using a neural multi-class occupancy function.
This approach utilizes a combination of occupancy networks~\cite{mescheder2019occupancy} and PointNet++~\cite{qi2017pointnet++} to infer a 3D object that aligns with the observation of the deformable real-world object.
It relies on deforming a pre-operative 3D model digitally to generate ground truth occupancy samples and a sensor-based depth image.
Through supervised learning, the network is trained to label occupancy samples correctly, learning the adaptation of the labelling to observed deformations directly from the data.
A distinct advantage of this method is its ability to estimate the position of deformable parts without necessitating initial registration.

To validate the accuracy of deformable registration and localization methods, organ "phantoms" are widely used, which aim to replicate the mechanical properties and medical imaging results of real organs.
Birnbaum et al.~\cite{birnbaum2007renal} customized renal inserts with different \gls{hu} values in anthropomorphic phantoms to produce varying brightness between kidney components under CT imaging.
However, the analysis of mechanical tissue properties was not undertaken.
In our previous work, we develop a hydrogel kidney phantom capable of reproducing both the mechanical and functional properties of living tissue~\cite{Ghazi_Kidney}.
This phantom is specifically designed for \gls{rapn} training, with a focus on mechanical fidelity over CT imaging characteristics.

A method that can register or reconstruct deformable objects from an observation, can be applied to perform resection tasks autonomously.
We recently presented the Autonomous System for Tumor Resection (ASTR)~\cite{ge2024autonomous} method.
It is a vision-guided robotic system that demonstrates success in tongue tumor resection (i.e. glossectomy).

In this work, we present an occupancy network-based tumor localization approach designed to use pre-operative CT images with intra-operative RGBD sensor data.
Our method only requires a single depth image from a single viewpoint to estimate the deformation of the kidney and locate embedded tumors at over $60\text{Hz}$, see \Cref{fig:figure_one}.
Therefore, it can track tumors, enabling the delineation of margins in the presence of deformations occurring during resection.
Furthermore, we evaluate our method through the use of a novel 3D hydrogel kidney phantom, which is embedded with renal tumors.
This phantom possesses realistic mechanical properties and allows easy automatic segmentation by providing varying brightnesses under CT imaging.
The evaluation demonstrates the utility of our method in guiding a robotic resection, accounting for potential deformations during partial nephrectomy procedures.

\section{Methods}

\newdimen\OccupancyNetworkX
\newdimen\OccupancyPointCloudY

\begin{figure*}
    \vspace{0.2cm} %
    \centering
    \resizebox{1\textwidth}{!}{
        \begin{tikzpicture}[node distance=0.3cm, auto]
          \begin{scope}[local bounding box=firstBox]
              \node (Phantom) [draw, rectangle, align=center] {\includegraphics[width=2cm]{graphics/main_figure/kidney_phantom_original.png}\\Phantom};
              \node (CT) [draw, rectangle, align=center, right=of Phantom] {\includegraphics[width=2cm]{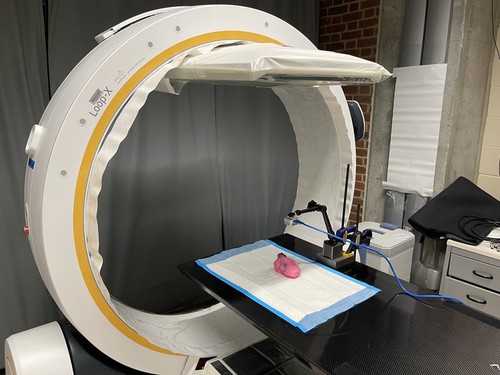}\\CT};
              \node (Volumetric Model) [draw, rectangle, align=center, right=of CT] {\includegraphics[angle=90, width=2cm]{graphics/main_figure/stacked_volumetric.png}\\Volumetric\\Model};
              \node (Surface Model) [draw, rectangle, align=center, right=of Volumetric Model] {\includegraphics[width=2cm]{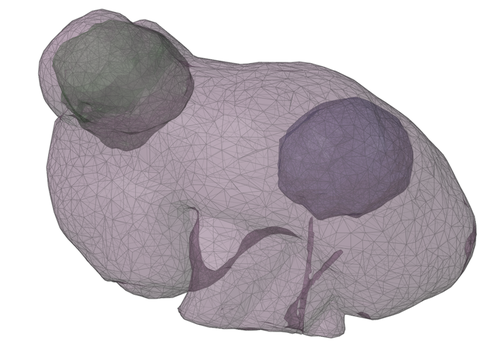}\\Surface Model};
              \node (Deformations) [draw, rectangle, align=center, right=of Surface Model] {\includegraphics[width=2cm]{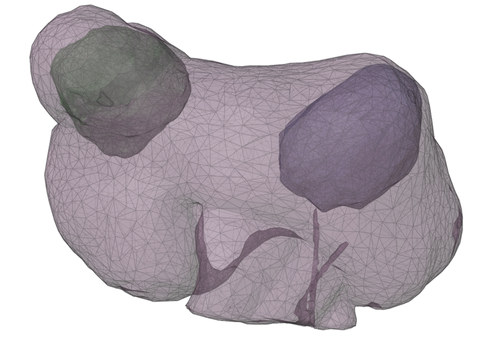}\\Apply\\Deformation};
              \node (Unlabelled Occupancy Cloud) [draw, rectangle, align=center, right=of Deformations] {\includegraphics[width=2cm]{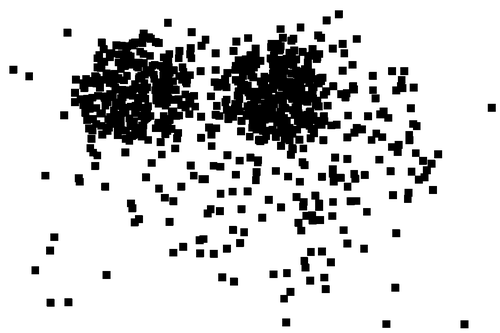}\\Unlabelled\\Occupancy Cloud};
              \node (Occupancy Network) [draw, rectangle, align=center, right=1.5cm of Unlabelled Occupancy Cloud] {\includegraphics[width=2cm]{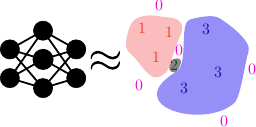}\\\textbf{Occupancy}\\\textbf{Network}};
              \pgfextractx{\OccupancyNetworkX}{\pgfpointanchor{Occupancy Network}{south}}

              \node (Sensor Point Cloud) [draw, rectangle, align=center, above=of Unlabelled Occupancy Cloud] {\includegraphics[width=2cm]{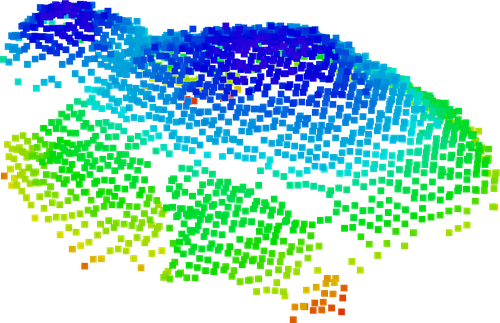}\\Sensor Point Cloud};
              \node (Labelled Occupancy Cloud) [draw, rectangle, align=center, below=of Unlabelled Occupancy Cloud] {\includegraphics[width=2cm]{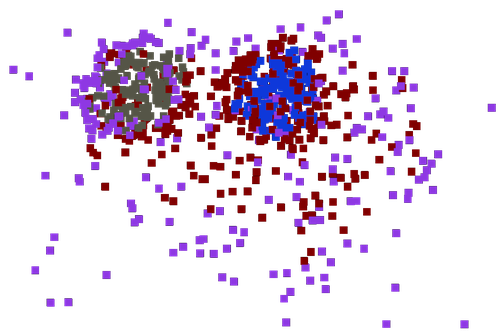}\\Labelled\\Occupancy\\Point Cloud};
              \pgfextracty{\OccupancyPointCloudY}{\pgfpointanchor{Labelled Occupancy Cloud}{east}}
              \node (Loss) [draw, rectangle, align=center, right=0.35cm of Labelled Occupancy Cloud]  {Loss};
              \node (Estimated Occupancy Labels)  [draw, rectangle, align=center] at (\OccupancyNetworkX, \OccupancyPointCloudY)  {\includegraphics[width=2cm]{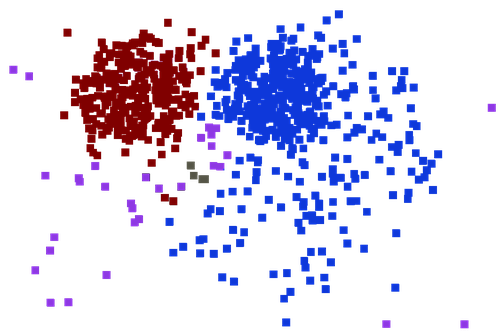}\\Estimated\\Occupancy\\Labels};
              
              \draw[->] (Phantom) -- (CT);
              \draw[->] (CT) -- (Volumetric Model);
              \draw[->] (Volumetric Model) -- (Surface Model);
              \draw[->] (Surface Model) -- (Deformations);
              \draw[->] (Deformations.north) to[bend left] (Sensor Point Cloud.west);
              \draw[->] (Sensor Point Cloud.east) to[bend left] (Occupancy Network.north);
              \draw[->] (Deformations.south) to[bend right] (Labelled Occupancy Cloud.west);
              \draw[->] (Labelled Occupancy Cloud) -- (Unlabelled Occupancy Cloud);
              \draw[->] (Labelled Occupancy Cloud) -- (Loss);
              \draw[->] (Loss) to[bend left] node[pos=0.5, sloped, above] {Update} (Occupancy Network);
              
              \draw[->] (Unlabelled Occupancy Cloud) -- (Occupancy Network);
              \draw[->] (Occupancy Network) -- (Estimated Occupancy Labels);
              \draw[->] (Estimated Occupancy Labels) -- (Loss);

              \node (Container) [draw, black, fit=(Deformations) (Estimated Occupancy Labels) (Sensor Point Cloud) (Labelled Occupancy Cloud) (Loss) (Occupancy Network) (Unlabelled Occupancy Cloud), inner sep=0.15cm] {};
              \node at (Container.north west) [below, inner sep=2mm, anchor=north west] {Training};

          \end{scope}
        
          \node[anchor=south west] at (firstBox.north west) {\textbf{Pre-Operative CT Imaging and Training}};
        
          \begin{scope}[yshift=-6.5cm] %
            \begin{scope}[local bounding box=secondBox]
                \node (Phantom) [draw, rectangle, align=center] {\includegraphics[width=2cm]{graphics/main_figure/kidney_phantom_original.png}\\Phantom};
                \node (Depth Image) [draw, rectangle, align=center, right=of Phantom] {\includegraphics[width=2cm]{graphics/main_figure/sensor_point_cloud.png}\\Sensor\\Point Cloud};
                \node (Occupancy Network) [draw, rectangle, align=center, right=of Depth Image] {\includegraphics[width=2cm]{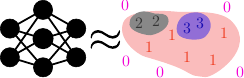}\\\textbf{Trained}\\\textbf{Occupancy}\\\textbf{Network}};
                \node (Random Query Points) [draw, rectangle, align=center , above=of Occupancy Network] {\includegraphics[width=2cm]{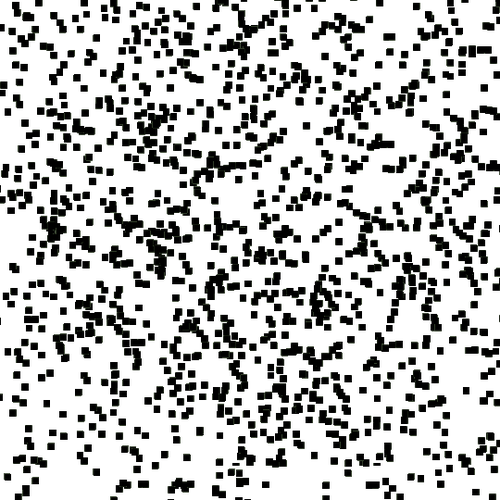}\\Random Query Points};
                \node (Output Occupancy Point Cloud) [draw, rectangle, align=center ,right=of Occupancy Network] {\includegraphics[width=2cm]{graphics/main_figure/output_occupancy_cloud_3d.png}\\Output Occupancy\\Point Cloud};
                \node (Robotic Resection) [draw, rectangle, align=center, right=of Output Occupancy Point Cloud] {\includegraphics[width=8.4cm]{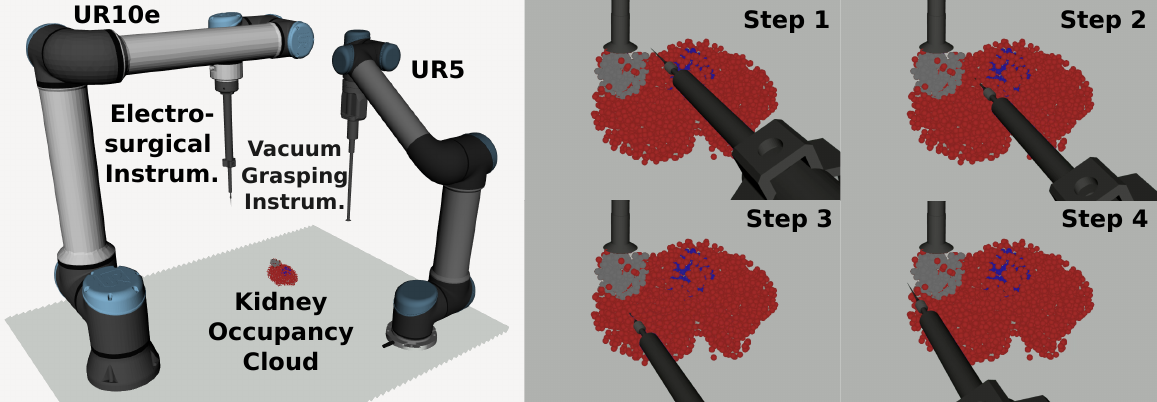}\\Robotic Resection};
                
                \draw[->] (Phantom) -- (Depth Image);
                \draw[->] (Depth Image) -- (Occupancy Network);
                \draw[->] (Random Query Points) -- (Occupancy Network);
                \draw[->] (Occupancy Network) -- (Output Occupancy Point Cloud);
                \draw[->] (Output Occupancy Point Cloud) -- (Robotic Resection);
            \end{scope}
            
            \node[anchor=south west] at (secondBox.north west) {\textbf{Intra-Operative Tumor Locazation and Robotic Resection}};
          \end{scope}
          
        \end{tikzpicture}
      }
    \caption{Our processing pipeline used to perform an interactive tumor localization and robotic resection in simulation.
    There are two processing blocks.
    The \textbf{Pre-Operative} part uses a model obtained from an imaging modality, such as a CT image, to produce training data for our occupancy network. During training, the occupancy network learns to estimate the locations and shapes of the tumors under deformations.
    The \textbf{Intra-Operative} part uses a sensor point cloud obtained from an RGBD camera to output an occupancy point cloud that is innately compatible with our surgical resection method.
    The visualizations for the occupancy networks are 2D simplifications.}
    \label{fig:main_figure}
\end{figure*}

    An overview of our method is shown in \Cref{fig:main_figure}.
    Here, we use an occupancy network which estimates the location of tumors based on a single-view point cloud of the kidney surface.
    The aim of our technique is to enable the localization to work in the presence of deformations.
    We make use of occupancy networks because they can directly encode deformable smaller structures inside of larger structures, such as a tumor inside of the kidney.
    This is in contrast to surface mesh based representations, which only provides a clear description of the surface of objects.
    Occupancy networks, therefore, directly provide 3D location and shape information of structures needed for autonomous resection.
    This innate suitability for robotic resection tasks is verified by integrating the occupancy-generated point cloud into an autonomous robot system, and virtually executing a resection plan using the robot.

    For this, a pre-operative model of the kidney with tumors is deformed virtually and used to train an occupancy network.
    In training, the occupancy network learns to estimate the 3D shape of the kidney based on a single depth image, re-projected into 3D space as a point cloud.
    We will refer to the point cloud obtained from the depth image as a \gls{dpp}.
    Conditioned on a real-world \gls{dpp}, the occupancy network produces registered volumetric information about the intra-operative kidney.
    The intra-operative localization of the tumors inside of a deformable kidney can be performed at over $60\text{Hz}$ on consumer hardware.
    This enables an interactive tracking of tumors during surgical applications.

\subsection{Localization}
\label{sec:localization}
    To locate the kidney tumors we use our deformable object reconstruction method as previously proposed~\cite{Henrich_2024_WACV}, with all losses and hyper-parameters remaining the same.
    To elaborate, PointNet++~\cite{qi2017pointnet++} is used as the encoder to distill a $1024$ dimensional latent representation.
    The latent vector is appended with a query point and passed to a fully connected \gls{mlp} with $8$ hidden layers of $512$ neurons.
    A skip connection is added to fifth hidden layer.
    ReLU is used as an activation function and batch normalization is used for the skip and the final layer.
    The network predicts the label for each query point using an argmax operation on the output logits, and the entire system is optimized end-to-end with a cross-entropy loss function.

    \noindent\textbf{Pre-Operative Training:}
    Our method requires training data to be created to perform the localization.
    For this, a 3D surface mesh of the kidney and the tumors is required.
    For example, these models can be obtained using pre-operative \gls{ct} images.
    An example for this is given in \Cref{fig:main_figure}.

    The occupancy network must learn how to estimate the tumor locations under deformation.
    Therefore, the training data needs to contain a deforming kidney.
    We propose a deformation system that ensures deformations are always partially visible to the camera.
    This ensures that reconstruction losses of parts not visible to the camera do not degrade the learning process.
    
    To produce a deformed kidney, we follow these steps:
    A set of all vertices on the kidney surface is created.
    This set is filtered, such that only vertices visible to the camera are contained.
    From this set, random vertices $v$ are moved by a random maximum distance, affecting each neighboring vertex $n$ according to a Gaussian falloff $N(|v-n|, \sigma^2)$.
    This includes vertices on the tumor meshes.
    The number of vertices and the distance that they should be moved is evaluated in \Cref{subsec:deformation_strength}.

    A \gls{dpp} is taken from the deformed model and SortSample~\cite{Henrich_2024_WACV} is used to obtain an occupancy point cloud.
    The occupancy network is then trained using the occupancy point cloud and the \gls{dpp} pairs.
    A total $31000$ data pairs is used for training and $1000$ are used for validation. 

    \noindent\textbf{Intra-Operative Inference:}
    For the robotic resection, see~\Cref{subsec:robotic_simulation}, a dense point cloud is needed which encodes volume and shape of the kidney and tumors.
    This dense point cloud is the same as an occupancy point cloud, without the outside points.
    To produce the needed dense point cloud, we query the trained occupancy network using $40000$ random query points.
    We discard all points that are not inside the kidney or the tumors.
    The occupancy point cloud is already registered to the camera and can be used without any modification as the input to our robotic resection method.

\subsection{Integration with Robotic Resection}
\label{subsec:robotic_simulation}

The integration of the registered occupancy point cloud into the robotic system is achieved using the Robot Operating System (ROS).
A robotic resection plan is implemented and executed virtually, visualized with the Robot Visualization Tool (RViz).
This process aims to verify the feasibility of using the dense occupancy point cloud to guide robotic resection, paving the way for future real-world applications.

We adapt ASTR \cite{ge2024autonomous} for autonomous nephrectomy.
As depicted in \Cref{fig:main_figure}, ASTR is equipped with an electrosurgical instrument on an UR10e manipulator (Universal Robots, Odense, Denmark), and a vacuum grasping tool on an UR5 manipulator.
The kidney's occupancy point cloud is streamed through ROS, and presented near ASTR as the surgical target.
A specialized planner in the ROS system directs the surgery step-by-step.
Initially, it localizes the exophytic tumor using labels and spatial data from the occupancy point cloud.
Subsequently, the tumor’s center of mass is calculated and mapped to the superior surface of its bounding box, delineating the target for the vacuum grasping tool, which then reaches in vertically and immobilizes the exophytic tumor for easier resection.
In the following step, the tumor point cloud is projected onto the transverse plane to identify a concave hull, delineating the tumor boundary.
This boundary is extended by $5\text{mm}$ for the resection margin, and the area of intersection with the kidney’s point cloud projected on the transverse plane is designated as the zone requiring cutting for tumor removal.
Points along the intersected contour are evenly spaced to guide the cutting process.
Currently, the planner employs a strategy of two sweeping cuts, mapping the intersected contour onto the central transverse plane of the tumor point cloud, and the posterior surface of its bounding box.
The electrosurgical instrument, positioned horizontally, follows these two paths to execute the cutting.
After completing the cuts, both robotic arms concurrently return to their initial poses.

\section{Experimental Evaluation}

\subsection{Experimental Setup}
\label{subsec:exp_setup}

\begin{figure}[ht]
    \centering
    \includegraphics[width=1\linewidth, trim={0cm 1cm 0cm 3cm}, clip]{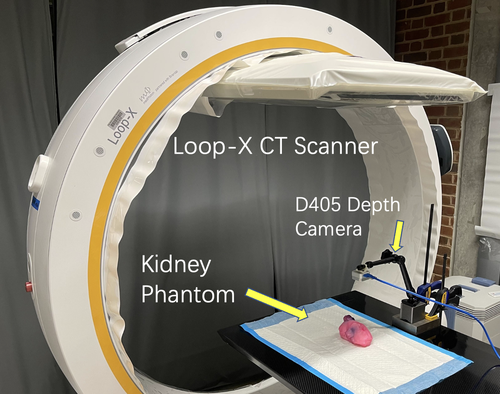}
    \caption{The kidney phantom is placed in the supine position onto the CT machine gantry. The RGBD camera is set up $15\text{cm}$ above the phantom.}
    \label{fig:experiment_setup}
\end{figure}

    The experimental setup is used to validate the tumor localization accuracy.
    A computed tomography (CT) machine (Loop-X, BrainLab, Germany) was used to scan the kidney phantom.
    An Intel RealSense D405 RGBD camera (Intel Corp., Santa Clara, California) captured the \gls{dpp} of the kidney phantom during the experiment, as shown in Fig.~\ref{fig:experiment_setup}.

    The CT scans were performed using a tube voltage of $90$ kV and an anode current of $13.6\text{mA}$, with a CT \gls{dap} of $500 \text{cGy}\cdot\text{cm}^2$ over $400$ slices.

    After each \gls{dpp} acquisition, and before CT scanning, the RGBD camera was moved out of the CT scanner's field of view to avoid artifacts from its metallic components.
    Care is taken not to touch or disturb the phantom and underlying supports during camera repositioning.
    This is to ensure the \gls{dpp}s correspond to the same deformed state captured in the CT scans.

\subsection{Kidney Phantom}
\label{subsec:kidney_phantom}

    \begin{figure*}
    \vspace{0.2cm}
    \centering
    \resizebox{1.0\textwidth}{!}{
        \begin{tikzpicture}[node distance=0.3cm, auto]
          \begin{scope}[local bounding box=firstBox]

               \node (InnerM) [rectangle, align=center] {$10\%$ PVA};
               \node (InnerT) [draw, rectangle, align=center, above=of InnerM] {Tumor + BaSO$_4$};
               \node (InnerB) [draw, rectangle, align=center, below=of InnerM] {Parenchyma};
               \node (Mixture) [draw, rectangle, fit=(InnerT) (InnerB) (InnerM), inner sep=0.1cm] {};

               \node (Phantom Mold) [draw, rectangle, align=center, left=of Mixture] {
                \includegraphics[width=2cm]{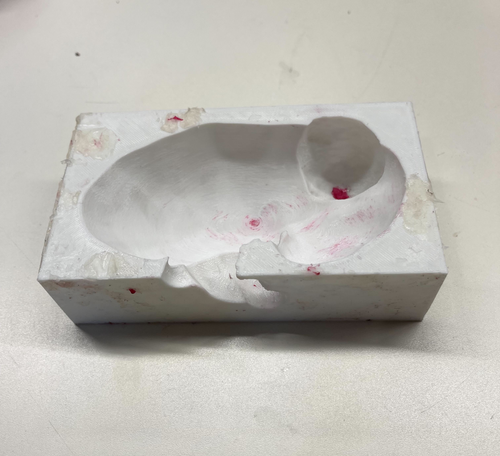}\\Phantom Mold
              };
              \node (Patient CAD) [draw, rectangle, align=center, left=of Phantom Mold] {
                \includegraphics[width=2cm,angle=90]{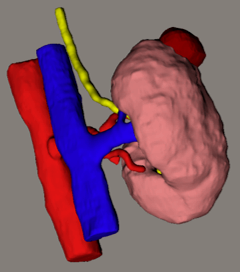}\\Patient CAD
              };
              \node (Segmentation) [draw, rectangle, align=center, left=of Patient CAD] {
                \includegraphics[width=2cm]{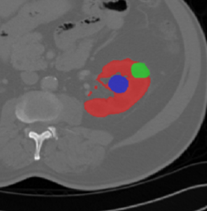}\\Segmentation
              };
              \node (Patient CT) [draw, rectangle, align=center, left=of Segmentation] {
                \includegraphics[width=2cm]{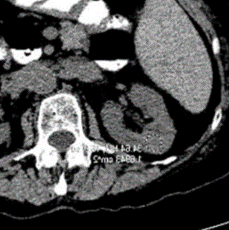}\\Patient CT
              };

              \node (Hydrogel Phantom) [draw, rectangle, align=center, right=of Mixture] {
                \includegraphics[width=2cm, angle=90]{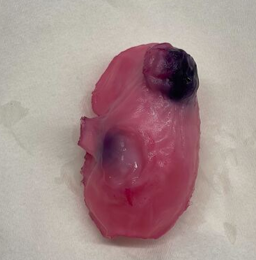}\\ Hydrogel Phantom
              };
              \node (CT Visualization) [draw, rectangle, align=center, right=of Hydrogel Phantom] {
                \includegraphics[width=2cm, angle=90]{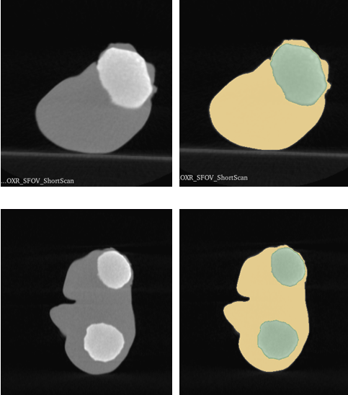}\\ CT Visualization
              };

              \draw[->] (Patient CT) -- (Segmentation);
              \draw[->] (Segmentation) -- (Patient CAD);
              \draw[->] (Patient CAD) -- (Phantom Mold);
              \draw[->] (Phantom Mold) -- (Mixture);
              \draw[->] (Mixture) -- (Hydrogel Phantom);
              \draw[->] (Hydrogel Phantom) -- (CT Visualization);

          \end{scope}
        \end{tikzpicture}
      }
    \caption{From real patient CT cases, we segmented the patient's kidney and tumors, generated a CAD model, and 3D printed the phantom mold. By injecting contrast-enhanced PVA liquid, we created a hydrogel phantom that exhibits clear tumor and parenchyma visualization under CT imaging.}
    \label{fig:workflow chart}
\end{figure*}
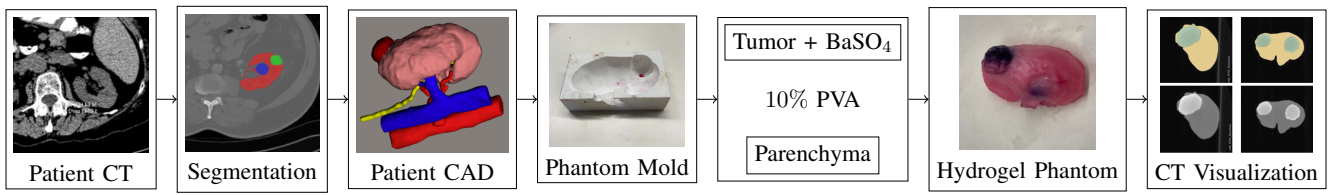

    We create two phantoms to fine-tune and evaluate the accuracy of our tumor localization method.
    The first phantom is used for the evaluation.
    It is created by combining two real-world patient cases into one phantom with a $38\text{mm}$ endophytic and $32\text{mm}$ exophytic tumor in one right kidney.
    The second phantom is used to fine-tune the amount of deformation that should be used during training, see \Cref{sec:localization}.
    This phantom is also based on a combination of two real-world patient cases.
    We use this second different phantom to ensure that we do not optimize our training data generation against our evaluation data.
    
    Two batches of 10\% polyvinyl alcohol (PVA) hydrogel were prepared, one with 2.27\% barium sulfate ($BaSO_4$, Sigma-Aldrich, St. Louis, MO) for contrast enhancement and the other without contrast agent.
    The contrast-enhanced PVA was injected into the tumor casts, while the plain PVA was used for the parenchyma cast.
    The exophytic tumor phantom was embedded directly into the cast.
    For the endophytic case, the tumor was suspended using wires before parenchyma injection.
    10\% PVA was then injected into the parenchyma casts, embedding the endophytic tumor.
    The entire phantom underwent freeze-thaw cycles to solidify and mimic mechanical tissue properties, with the final cycle facilitating cohesion between structures to replicate connective tissues.
    
    The experiment aimed to simulate the in vivo kidney deformation that occurs when the kidney is subjected to a parenchymal incision under controlled renal arterial pressure during partial nephrectomy.
    Altamar et al. developed a kidney elastic deformation model describing the deformation caused by incision and loss of turgor \cite{Altamar_Deformation}.
    The model indicates a general "sinking in" of the kidney surface due to gravity, accompanied by a small expansion along the lateral edges.
    In this biomechanical model, the maximum deformation is the indentation at the midsection of the kidney, which was approximately $8\text{mm}$. The average displacement of the lateral edges is around $4.4\text{mm}$.
    
    Based on this elastic model, we force the kidney phantoms to deform.
    To simulate the "sinking in" surface configuration with expansion along the edges while keeping the surface relatively smooth without sudden indentations, we use a soft cardboard adhered to the anterior and posterior surfaces of the kidney phantom, see Fig.~\ref{fig:Deformation_results}(a) and (b).
    Nylon cable ties are punctured through the soft cardboard and phantoms, locking on the other side.
    By adjusting the tightness of the cable ties, varying degrees of compression can be applied to the cardboard, and subsequently to the phantoms.

\subsection{Experiments}
\label{sec:experiments}

    To evaluate the accuracy of our tumor localization method, we use the first kidney phantom described in \cref{subsec:kidney_phantom}.

    Through deformation and rotation, we create 6 kidney variants from our original phantom.
    Three variants contain increasing deformations and three contain different orientations.

    \begin{figure}[h]
    \centering
    \includegraphics[width=1\linewidth, trim={0.4cm, 0.4cm, 0.4cm, 1.2cm}, clip]{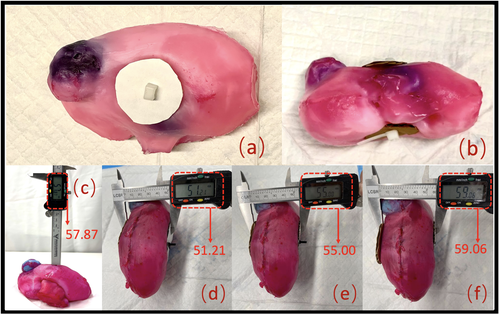}
    \caption{Figure (a),(b) and (c) depict the kidney phantom in a normal supine configuration with a midsection height of approximately $57.87\text{mm}$. Figures (d), (e) and (f) present the measurements of the midsection under compression, which are $51.21\text{mm}$, $55.00\text{mm}$, $59.06\text{mm}$}
    \label{fig:Deformation_results}
\end{figure}

    As described in \Cref{subsec:kidney_phantom}, we generate three different degrees of deformations with midsection compressed by $4\text{mm}$, $8\text{mm}$, and $12\text{mm}$, see \Cref{fig:Deformation_results}.
    We denote them as \textit{D1}, \textit{D2} and \textit{D3}, respectively.
    The three orientations are generated by rotating the kidney phantom by 90, 180 and 270 degrees.
    We denote them as \textit{Rot90}, \textit{Rot180} and \textit{Rot270}.
    For each kidney variant, the experimental setup described in \cref{subsec:exp_setup} is used to capture a CT image and a \gls{dpp}.

    The occupancy network requires the \gls{dpp} to only contain the kidney.
    Therefore, we discard points further than $15\text{cm}$ from the camera.
    We also discard points closer than $11\text{cm}$ to remove some floating artifacts.
    We scale up the \gls{dpp} by a factor of $1000$, to match the scaling of the CT.
    1000 unique points are randomly sampled from the \gls{dpp}, to be used as the input to the point cloud encoder~\cite{Henrich_2024_WACV}.
    
    All meshes required for training are automatically extracted from the CT data.
    For this, the kidney and tumors are segmented using an upper-lower threshold segmentation, with the value ranges being $(-300, 300)$ and $(300,1000)$~\gls{hu}, respectively.
    The segmentation masks are converted to surface meshes.
    To remove small floating artifacts, only large connected triangle surfaces are kept.

    The occupancy network is trained using training data created from the original kidney model.
    The kidney's relative position and rotation to the camera may be unknown.
    We randomly rotate the original kidney virtually.
    As the kidney is round with few features, estimating the rotation is difficult from a single point cloud.
    We therefore produce two datasets.
    The first rotates the kidney randomly by up to $360$ degrees around all axes.
    The second rotates the kidney by up to $360$ degrees around the Z axis, but the random rotations for X and Y are uniformly sampled from the interval $[-15, 15]$ (degrees).
    The Z axis points upwards in \Cref{fig:experiment_setup}.
    The model trained on the fully random rotations is denoted as $M_{\text{full}}$, the other is $M_{\text{lim}}$, both of which are evaluated.
    
    The kidney variants are used to obtain the evaluation data.
    For this, the point cloud and the meshes of the variants are manually registered.
    A rigid object placed next to the CT is used as a registration aid.

    The point clouds of the kidney variants are then passed to the trained occupancy network.
    The occupancy network is sampled on an equidistant grid with a resolution of $400 \times 400 \times 400$ to produce a volumetric model.
    Although our robotic application does not require such a dense volumetric representation, it enables a more accurate evaluation.
    The volumetric model is then compared to the reference kidney models to compute the \gls{hd}.
    In the context of estimated and reference tumors, the \gls{hd} quantifies the maximum distance from any point on the estimated tumor surface to the nearest point on the real tumor surface.
    Consequently, this measurement provides an estimate for safety margin that should be added into the surgical excision of the tumor to ensure that it is completely resected.
    Additionally, we compute the center of the axis aligned bounding boxes for both the volumetric model and the reference object.
    The distance between these centers are computed, to evaluate how well the positioning of the tumors was estimated.

    We also evaluate how varying the degree of deformation during training affects the accuracy of the output.
    For this, the second phantom is used.
    This evaluation is performed using limited rotations, like for $M_{\text{lim}}$.
    We apply a varying number of successive vertex-pull operations, as described in \Cref{sec:localization}.
    We use \gls{miou} as a single metric that describes how well the kidney with all integrated tumors is estimated.

\section{Results}

    \subsection{Training and Inference Time}
    All measurements are peformed using a single Nvidia RTX 4090 GPU (Nvidia, Santa Clara, California).
    During training, each model achieves it's maximum validation accuracy after approximately $2$ hours or $200$ epochs.
    At inference time, the time needed to label $40000$ query points, with a \gls{dpp} containing $1000$ points, is approximately $14\text{ms}$.

    \subsection{Deformation Strength}
        \label{subsec:deformation_strength}

\begin{table}[h]
\centering
\caption{Effect of using different deformations configurations on mIoU. Each \textit{(Distance, $\sigma^2$, Times Repeated)} triple represents a vertex-drag operation. Where a random vertex $v$ is moved according to \Cref{sec:localization}.}
\label{tab:compare_deformations}
\begin{tabularx}{\columnwidth}{rXc}
\hline
\textbf{Conf.} & \textbf{Deformations (Distance, $\sigma^2$, Times Repeated)} & mIoU\\
\hline
1 & (16, 33.3, 1) & 0.70\\
\hline
2 & (16, 33.3, 1), (16, 16.7, 1), (8, 11.7, 1) & 0.65\\
\hline
3 & \begin{tabular}[c]{@{}l@{}}(8, 11.7, 1), (6, 10, 1), (4, 6.7, 1), (4, 3.3, 5),\\(2, 3.3, 10)\end{tabular} & \textbf{0.71}\\
\hline
4 & (16, 13.3, 1), (16, 10, 1), (16, 6.7, 1) & \textbf{0.71}\\
\hline
5 & (10, 33.3, 1), (16, 13.3, 1), (16, 6.7, 1) & 0.68\\
\hline
6 & \begin{tabular}[c]{@{}l@{}}(16, 33.3, 1), (16, 16.7, 1), (8, 11.7, 1), (6, 10, 1),\\(4, 6.7, 1), (4, 3.3, 5), (2, 3.3, 10)\end{tabular} & 0.66\\
\hline
\end{tabularx}
\end{table}

        Six different deformation configurations are tested, see \Cref{tab:compare_deformations}.
        Overall, configuration $3$ and $4$ performed best, each with an \gls{miou} of $0.71$.
        For all following evaluations, we use deformations as defined in configuration $3$ for creating training data.

    \subsection{Localization Accuracy}
    \label{subsec:localization_accuracy}
        \input{tables/results}
    
        We evaluate the accuracy of the occupancy networks $M_{\text{full}}$ and $M_{\text{lim}}$.
        Trained on fully and limited rotations respectively, see \Cref{sec:experiments}.
        The results are shown in \Cref{tab:results_reconstruction}.
        For scenes with small deformation, such as \textit{D1}, $M_{\text{full}}$ and $M_{\text{lim}}$ perform similarly.
        Both achieve a \gls{hd} of approximately $4\text{mm}$ and $7\text{mm}$ for the exophytic and endophytic renal tumors, respectively.
        For \textit{D2}, both models again perform similarly, with tumor \gls{hd}s of approximately $6\text{mm}$ and $9\text{mm}$.
        For the strongest deformation \textit{D3} we evaluate, $M_{\text{lim}}$ better estimates the tumors, with \gls{hd}s of $11.3\text{mm}$ and $12.9\text{mm}$ versus $15.4\text{mm}$ and $18.5\text{mm}$.
        For the evaluated variants \textit{Rot90}-\textit{Rot270}, $M_{\text{full}}$ estimates both tumors with \gls{hd}s of less than or equal to $13.4\text{mm}$.
        For all kidney and model combinations, except \textit{D3} with $M_{\text{full}}$, the tumor centers are estimated with an error of less than $10\text{mm}$.

\section{Discussion and Conclusion}

    For scenes with little deformations, our models trained on limited rotations $M_\text{lim}$, and trained on full rotations $M_\text{full}$ perform similarly.
    However, as deformations get stronger (\textit{D3}), $M_\text{full}$ is unable to accurately reconstruct the tumors.
    This is likely due to orientations being difficult to distinguish in the presence of strong deformations.
    Conversely, $M_\text{lim}$ demonstrates a relatively consistent ability to estimate the tumor positions across different deformations.
    As it has only seen a limited amount of rotations during training, it has more inductive bias.
    It is able to make more assumptions about the input data when making predictions.
    During minimally invasive surgery, the view of the kidney is limited, we therefore suggest limiting the rotations accordingly.

    Generally, the \gls{hd} of the exophytic tumor, compared to the endophytic tumor, is lower.
    The exophytic tumor is often, at least partially, visible to the depth sensor, providing features for estimating the location.

    The \gls{hd} values for both tumors suggest a safety margin of approximately $16\text{mm}$ to $19\text{mm}$ around the exophytic and endophytic tumors respectively.
    For smaller kidney deformations, safety margins of $6\text{mm}$ and $10\text{mm}$ are sufficient.

    We plan to transfer our simulated resection to the real-world to perform a tumor resection on our phantom.
    For this, we must take visual obstructions into account.
    Further, we aim to refine the dual-arm resection technique, advancing from the current basic two-sweep strategy.
    This will include active tumor retraction through vacuum grasping, and finer forward movements at diverse angles with the electrosurgical instrument during a single sweep.

    \noindent\textbf{Method Limitations:}
    Our results indicate that our current approach struggles to capture local deformations on the surface of the kidneys.
    For this, we will further improve our deformation system, used to produce training data.
    Adding regularizers, such as volume-retention, may improve the tumor locations estimation.

    \noindent\textbf{Evaluation Limitations:}
    When registering the \gls{dpp} and the CT images, we noticed distortions and artifacts in the \gls{dpp}.
    This prevents an optimal registration, directly affecting our results.
    The move away from our low cost depth camera to a more accurate model can resolve this.
    The cardboard and cable tie method for registration causes the \gls{dpp} to have a systematic offset due to the thickness of the cardboard.
    Using sutures to deform the kidney could resolve this issue.
    Finally, our manual registration introduces an error.
    Even with registration guides, we notice that our \gls{dpp} and CT images do not align correctly.
    Again, this may be resolved with a better depth sensor.
    Using a virtual camera, to produce the \gls{dpp} directly from the CT images, could provide a quantitative measure for the error caused by our depth sensor.

    \noindent\textbf{Conclusion:} We present and evaluate an occupancy network based method to determine the location of exophytic and endophytic kidney tumors in a deformable phantom.
    This method only requires a single depth image.
    The accuracy of the location estimate varies with the strength of the kidney deformation.
    As the deformation increases, a larger safety margin must be chosen.
    For smaller kidney deformations, a safety margins of $10\text{mm}$ and $12\text{mm}$ for the exophytic and endophytic tumors are needed.
    This estimation can be performed over $60$ times a second.
    Further, we show that the use of an occupancy network directly enables down-stream tasks, such as a robotic resection.
    Currently, multiple limitations affect our evaluation and method accuracy.
    We discuss their possible solutions.

\bibliographystyle{IEEEtran.bst}
\bibliography{iros.bib}

\end{document}